\documentclass[10pt,twocolumn,letterpaper]{article}

\usepackage{wacv}
\usepackage{times}
\usepackage{epsfig}
\usepackage{graphicx}
\usepackage[fleqn]{amsmath}
\usepackage{amssymb}
\usepackage[utf8]{inputenc}
\usepackage{tabularx}
\usepackage{wrapfig}
\usepackage{multirow}
\usepackage{natbib}
\usepackage[table]{xcolor}
\usepackage{listings}

\usepackage[pagebackref=true,breaklinks=true,letterpaper=true,colorlinks,bookmarks=false]{hyperref}

\wacvfinalcopy 

\newcolumntype{C}[1]{>{\centering}m{#1}}
\newcolumntype{L}{>{\centering\arraybackslash}m{0.7cm}}

\def\wacvPaperID{958} 

\ifwacvfinal\fi
\setcitestyle{numbers}
\begin{document}

\title{Rainy screens: Collecting rainy datasets, indoors.}

\author{Horia Porav \\
Oxford Robotics Institute,\\ University of Oxford\\
{\tt\small horia@robots.ox.ac.uk}
\and
Valentina-Nicoleta Musat \\
Oxford Brookes University\\
{\tt\small 16075442@brookes.ac.uk}
\and
Tom Bruls \\
Oxford Robotics Institute,\\ University of Oxford\\
{\tt\small tombruls@robots.ox.ac.uk}
\and
Paul Newman \\
Oxford Robotics Institute,\\ University of Oxford\\
{\tt\small pnewman@robots.ox.ac.uk}
}

\maketitle
\ifwacvfinal\thispagestyle{empty}\fi

\begin{abstract}
   Acquisition of data with adverse conditions in robotics is a cumbersome task due to the difficulty in guaranteeing proper ground truth and synchronising with desired weather conditions. In this paper, we present a simple method - recording a high resolution screen - for generating diverse rainy images from existing clear ground-truth images that is domain- and source-agnostic, simple and scales up. This setup allows us to leverage the diversity of existing datasets with auxiliary task ground-truth data, such as semantic segmentation, object positions etc. We generate rainy images with real adherent droplets and rain streaks based on Cityscapes and BDD, and train a de-raining model. We present quantitative results for image reconstruction and semantic segmentation, and qualitative results for an out-of-sample domain, showing that models trained with our data generalize well. 
\end{abstract}

\section{Introduction}
Bad weather conditions are among the biggest challenges in robotics. Fog, snow, haze, rain and dirt can affect both the visual perception of humans and severely diminish the performance of computer vision tasks.

The effects of impurities on visual perception are intricate, but can be classified in two main categories. Firstly, by impurities in the atmosphere that occlude the scene (snow, rain streaks, fog, smog) but which have no distorting effects. Secondly, by impurities that adhere to transparent surfaces and have distorting effects. For example, adherent rain drops act as a fish-eye-lens due to their convex shape, with the image formed inside the droplet being a result of light rays coming from an area that is larger than what the droplet is occluding. 


Modelling adherent droplets is a difficult task due to the various sizes and shapes that they can take, distortion, zoom, blur, glare, the effect of gravity, and cohesion and adhesion of water \cite{you2016b}. Moreover, although the scene is in focus, the images formed inside droplets in front of the camera is usually blurry and opaque. Finally, if the light refracted inside the droplet comes from a large part of the scene, the image formed in the droplet becomes gray \cite{ulfwi2018}. 


\begin{figure}[!tbp]
\centering
{\includegraphics[width=0.496\columnwidth]{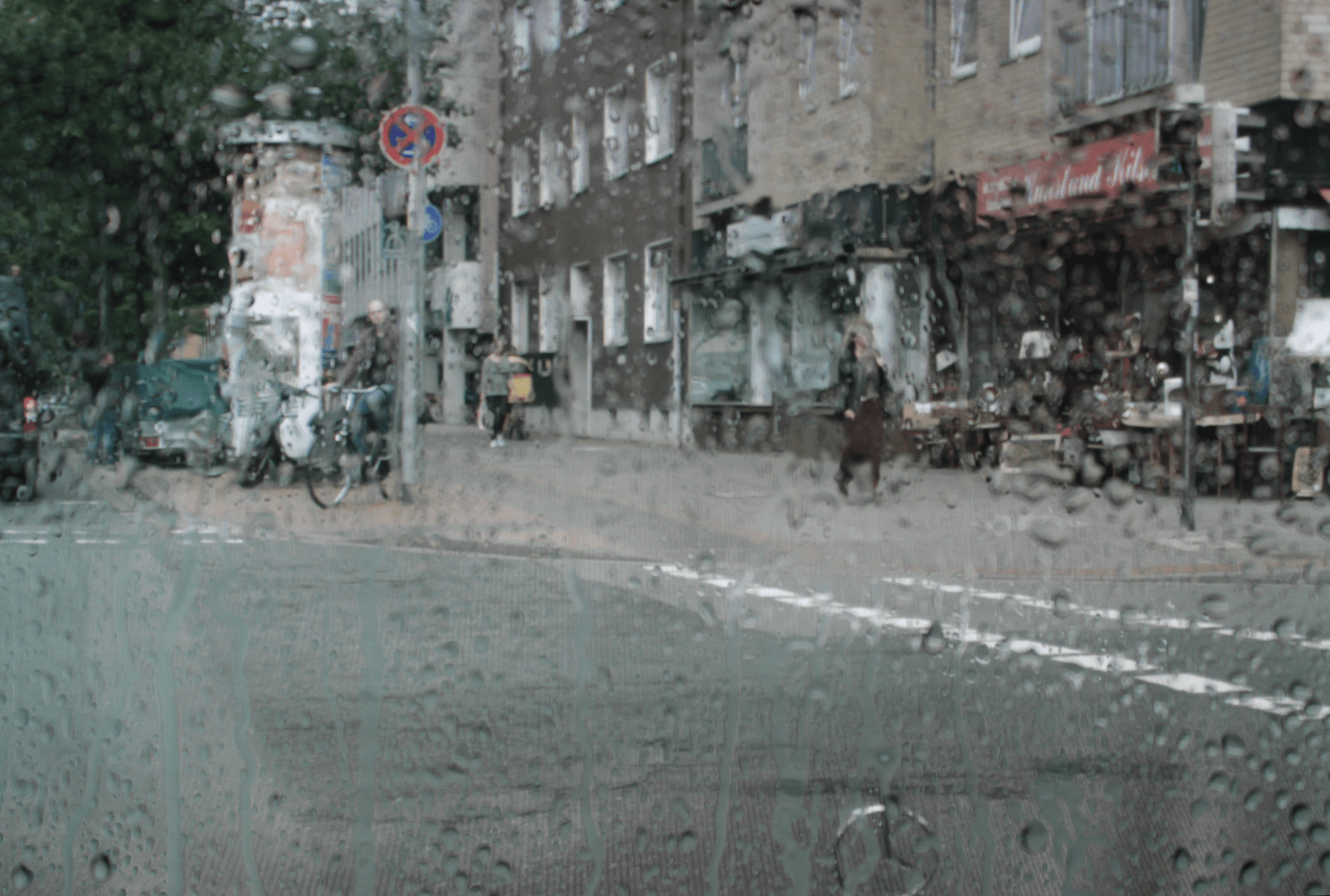}}%
\hfill 
{\includegraphics[width=0.496\columnwidth]{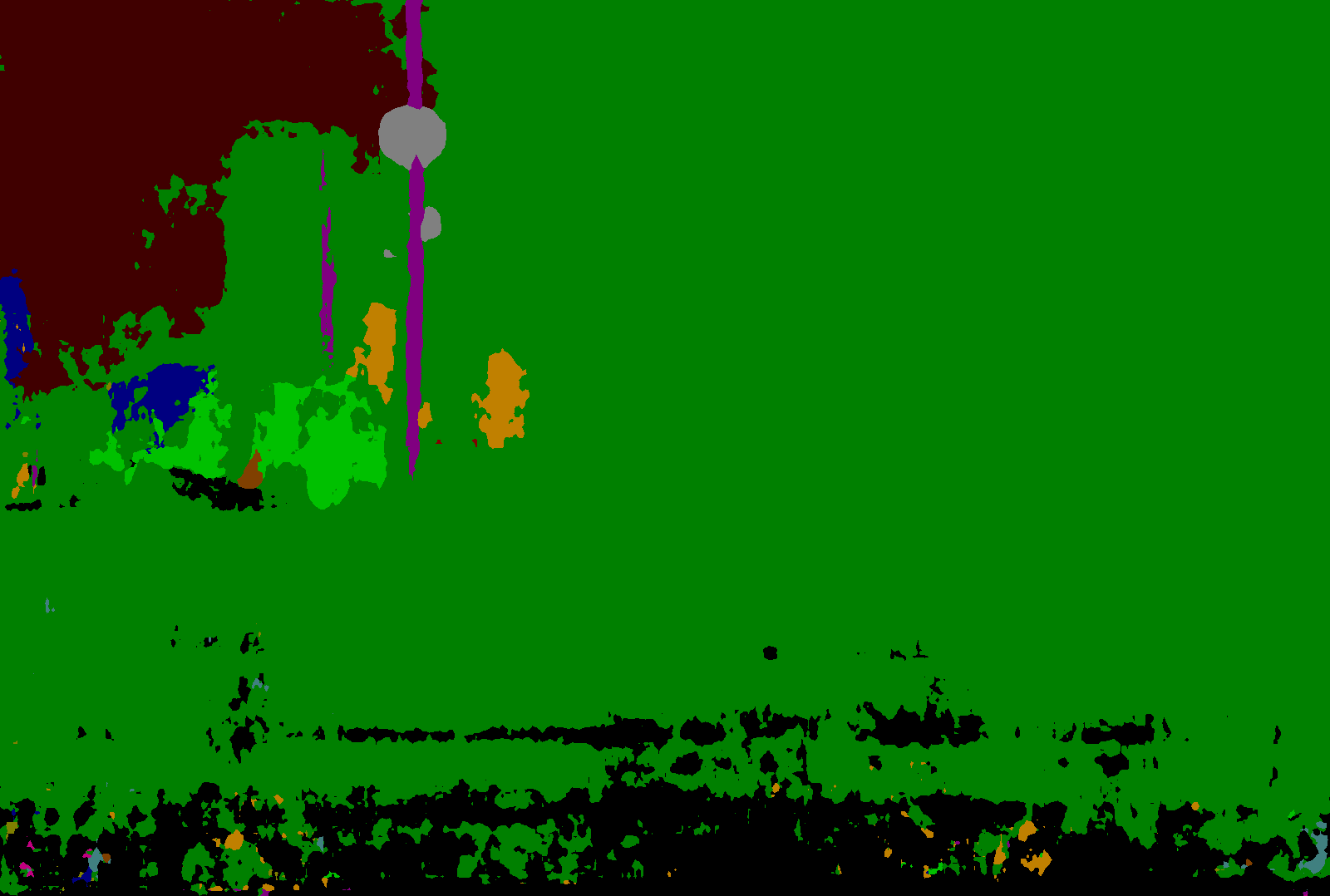}}%
\\ 
\vspace{0.2mm}
{\includegraphics[width=0.496\columnwidth]{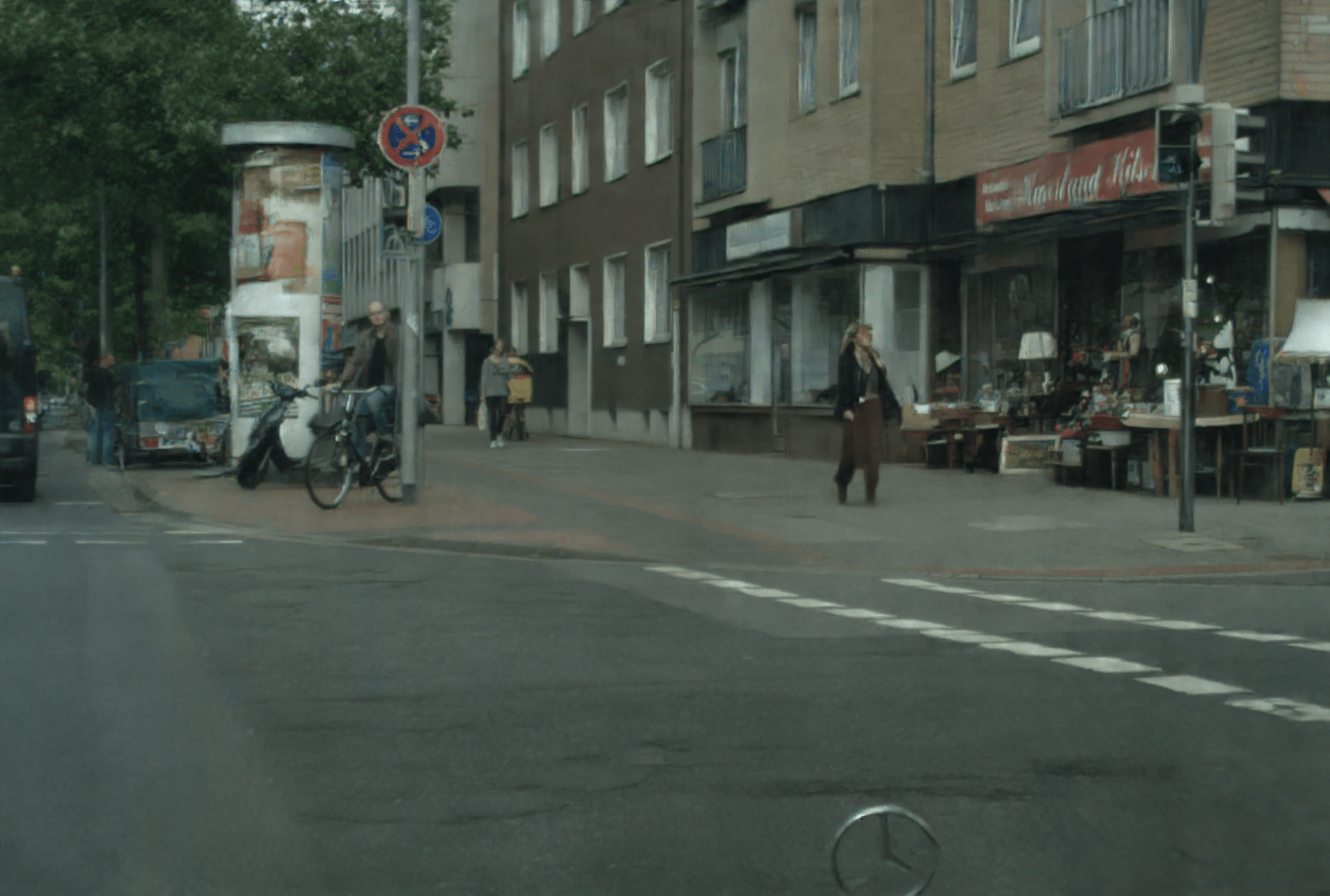}}%
\hfill 
{\includegraphics[width=0.496\columnwidth]{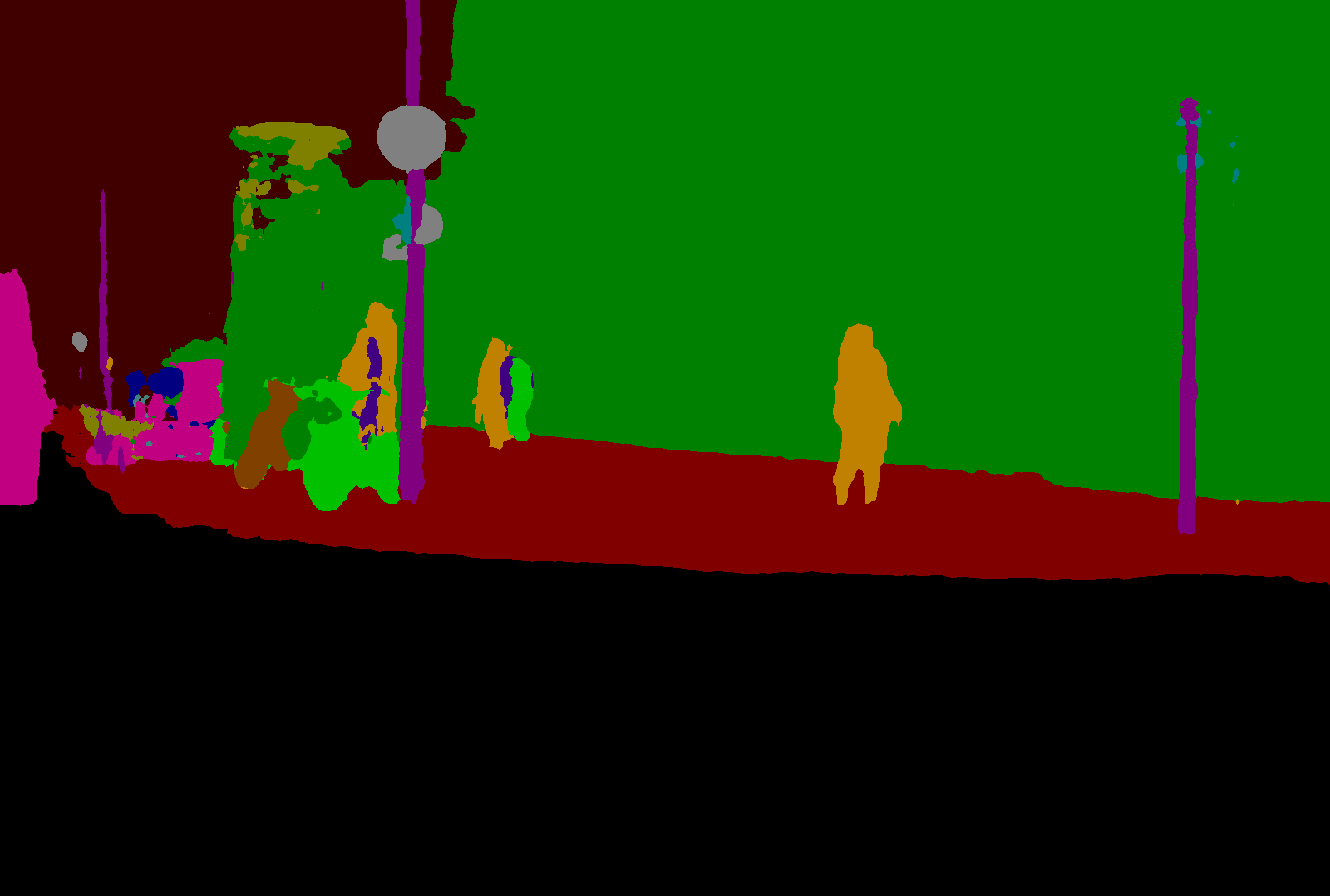}}%
\caption{Top row: rainy image and its semantic segmentation. Bottom row: derained image and its semantic segmentation. The rainy image was obtained from a clear, dry-condition dataset (Cityscapes), using an indoor, automated data collection rig consisting of a high-resolution monitor, a DSLR camera, glass panes and water spray nozzles. This setup facilitates the creation of high-quality rainy datasets complete with a perfectly aligned clear, dry groundtruth and \textbf{any} other attributes of the original dataset such as semantic segmentation labels, depth etc. Once set up, our method provides automatic, unsupervised image capture, with a wide variety of water droplet sizes and streak patterns.}
\label{taster}
\end{figure}

To the best of our knowledge, there are three main methods through which rainy images with ground-truth are generated in existing literature. The first is manual, and involves using a glass pane to capture a clean and a rainy image sequentially \cite{qian2017}. While this offers perfect alignment, it is tedious, and does not scale up to dynamic scenes due to motion and changes in illumination. The second method involves using stereo cameras, where one lens is kept clean and one rainy, with both images captured simultaneously \cite{porav2019}. Although the scenes are more diverse and the method scales up, it assumes that the two images can be aligned using a homography which, unfortunately, only works on planar scenes. The third method involves computer-generated rain, which scales up, but there is a large domain gap between real rain and computer-generated rain \cite{porav2019}. 

Building datasets with both real rain and clear ground truth under hard-to-control outdoor environments is very difficult for static images, let alone scenes containing moving objects. One has to account for changes in illumination, motion, occlusion/dis-occlusion, changes in appearance, and so on. For this reason, many of the datasets focus on static scenes, while those with video sequences do not offer non-rainy ground truth, apart from \cite{porav2019}.

We propose a simple method for capturing diverse raindrops, that can make use of existing datasets with clear images and auxiliary task ground truth, but also of arbitrary footage. Our method is inspired by \cite{porav2019}, which use a water pump to spray water on a glass pane, and \cite{hamileh}, which use a windshield plane that can be tilted and translated in front of the camera. The difference is that we capture monocular images by recording a high-resolution screen.
The advantages of this approach are threefold:
\begin{itemize}
    \item perfectly aligned image pairs;
    \item high scalability, with no requirement for a mobile platform such as a vehicle since all data collection happens indoors, unsupervised, and in a controlled environment; and
    \item diverse, in and out of focus droplets and streaks.
\end{itemize}
In this way, we are able to easily generate image pairs with high diversity, and further demonstrate that data gathered indoors using this method can efficiently be used to both train and pre-train deraining models that generalize well.
\section{Related work}
\noindent\textbf{Rain modelling:} A photometric raindrop model is used by \cite{hamileh} and \cite{roser2009} to construct the appearance of raindrops by tracing the light rays that pass through them, with the droplet boundary modelled by a sphere with equal contact angles. In a follow-up paper \cite{roser2010}, 2D Bezier curves were used to incorporate raindrop deformation due to the effect of gravity. Later in \cite{you2016a}, the dark band around them is produced with the total reflection phenomenon (light from the environment being reflected back into the waterdrop). Based on \cite{hamileh}, \cite{roser2009} and \cite{you2016a}, the authors of \cite{porav2019} propose an approach similar to meta-balls \cite{blinn1982}, where they model synthetic waterdrops by warping normal maps of droplets.
\vspace{1mm}
\newline
\noindent\textbf{Adding droplets:} In the experimental setup of \cite{eigen2013}, rain is simulated by spraying water on a glass pane in front of the camera, but due to changes in illumination and subject motion, no ground truth is provided. Additionally, a video sequence with real rain is provided, without ground truth. The authors of \cite{you2016b} run experiments to study the effects of disturbance of light sources in environment, various shapes and sizes of raindrops, blurred raindrops and glare, in a droplet detection and removal pipeline in videos. They use a glass pane on which they spray water and capture videos in the real world, however ground truth is only available for the droplets' position. 

Complete reconstruction ground truth (clean and rainy image pairs) was first attempted in \cite{qian2017}. To ensure alignment and keep the same refractive index, they use both a sprayed glass pane and a clean glass pane. Additionally, they ensure that the atmospheric conditions and background remain constant during the two shots. They provide 1119 image pairs, however the process is difficult to scale up, and since background needs to be kept constant between the two shots, the images contain only static scenes.

On the other hand, the authors of \cite{porav2019} use a bi-partite chamber that is placed in front of a small-baseline stereo camera, keeping one part dry while the other part is sprayed with water. The system was mounted on a vehicle and image sequences were taken while driving. The authors provide 50000 pairs of undistorted, cropped and aligned images with both rain streaks and raindrops of various shapes and sizes. Although the system is portable and image acquisition is simpler, the method requires a mobile platform, and the alignment is based on a homography that assumes the world is flat, resulting in a small degree of misalignment between image pairs.
\vspace{1mm}
\newline
\noindent\textbf{Raindrop detection and removal:} The classical approaches for detecting and removing raindrops rely on a combination of image segmentation, pattern recognition, ellipse fitting, template matching \cite{ito2015}, \cite{vijay2018}, \cite{fouad2019}, \cite{roser2009}, \cite{roser2010}, multiple image analysis \cite{you2016b}, or multiple and/or rotating cameras \cite{kuramoto2002}, \cite{yamashita2003}, \cite{yamashita2004} and \cite{yamashita2005}. More recent approaches employ Convolutional Neural Networks and/or adversarial frameworks \cite{eigen2013}, \cite{qian2017}, \cite{lin2018}, while in \cite{ulfwi2018} both classical methods and CNNs are combined.
\vspace{1mm}
\newline
\noindent\textbf{Soil and dirt removal:} Since cameras mounted outside the vehicle are exposed to adverse conditions, fully autonomous cars need a system to detect mud and trigger a cleaning mechanism. Recent work done in this area is that of \cite{uricar2019}, which create a dataset with opaque and transparent soiling used to train a soiling detection network. However, since the acquisition of realistic muddy images is a difficult task, they propose a GAN-based architecture to generate artificial soil on clean images. Similarly, the authors of \cite{eigen2013} generate synthetic dirt for training purposes, and validate their results on both synthetic dirt, and real dirt captured using a glass pane. The authors of \cite{michaelis2019} study how object detection is affected when degrading the image quality, by introducing synthetic fog, frost, snow and other corruptions and effects.
\section{Methodology}
\subsection{Data acquisition}
\begin{figure}[!htbp]
\centering
{\includegraphics[width=\columnwidth]{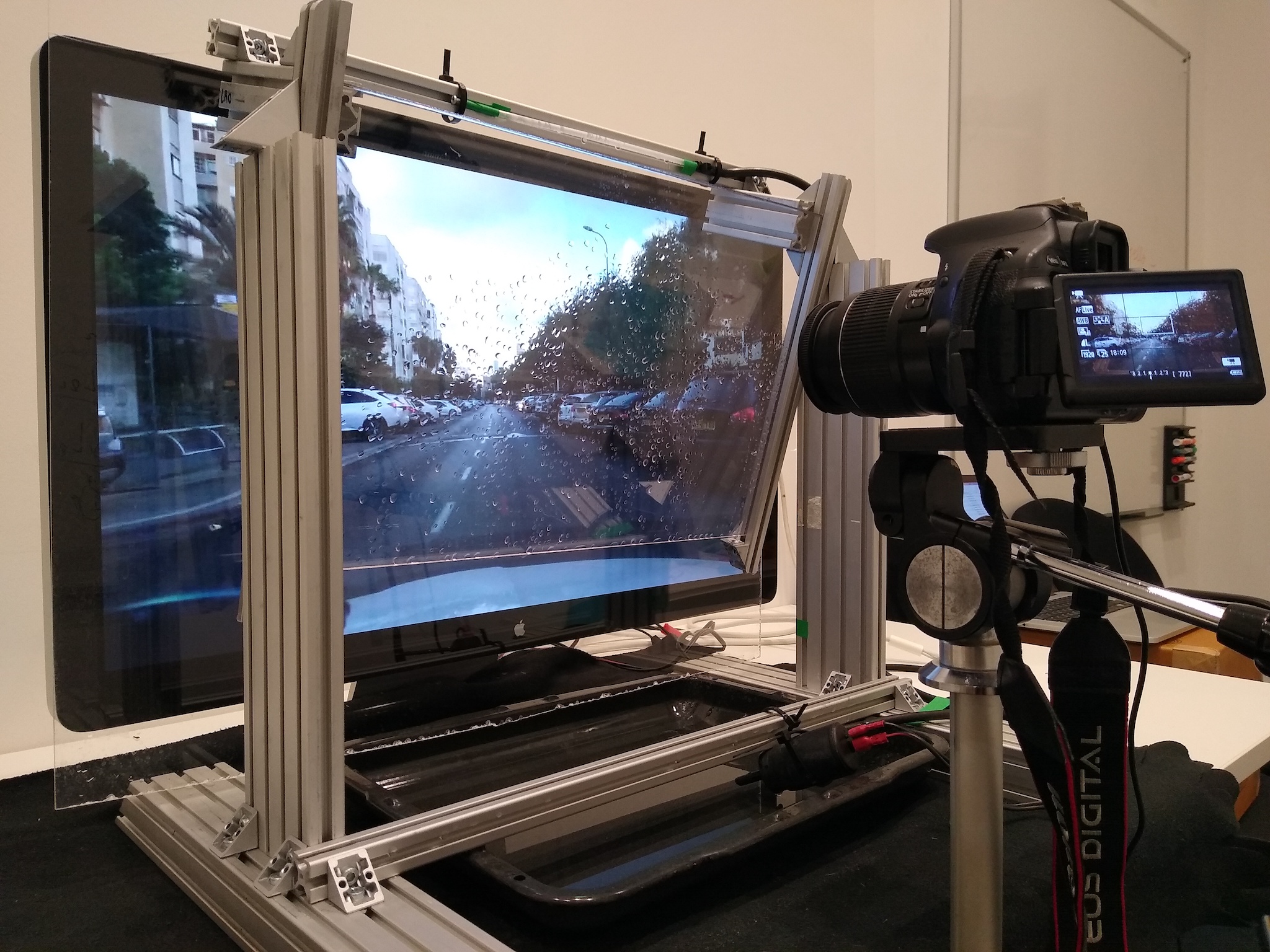}}%
\caption{Rain-maker camera rig. Note that the datasets are collected with the room lights OFF!}
\label{fig:rig}
\end{figure}

We build a rig consisting of a camera on a tripod, a tiltable glass pane simulating a car windshield, multiple water spray nozzles with a recirculating pump and collector tray, and a waterproofed high-resolution (2560x1440) computer monitor. The setup is presented in figure Fig \ref{fig:rig}.

\begin{figure*}[!htbp]
\centering
{\includegraphics[width=0.33\textwidth]{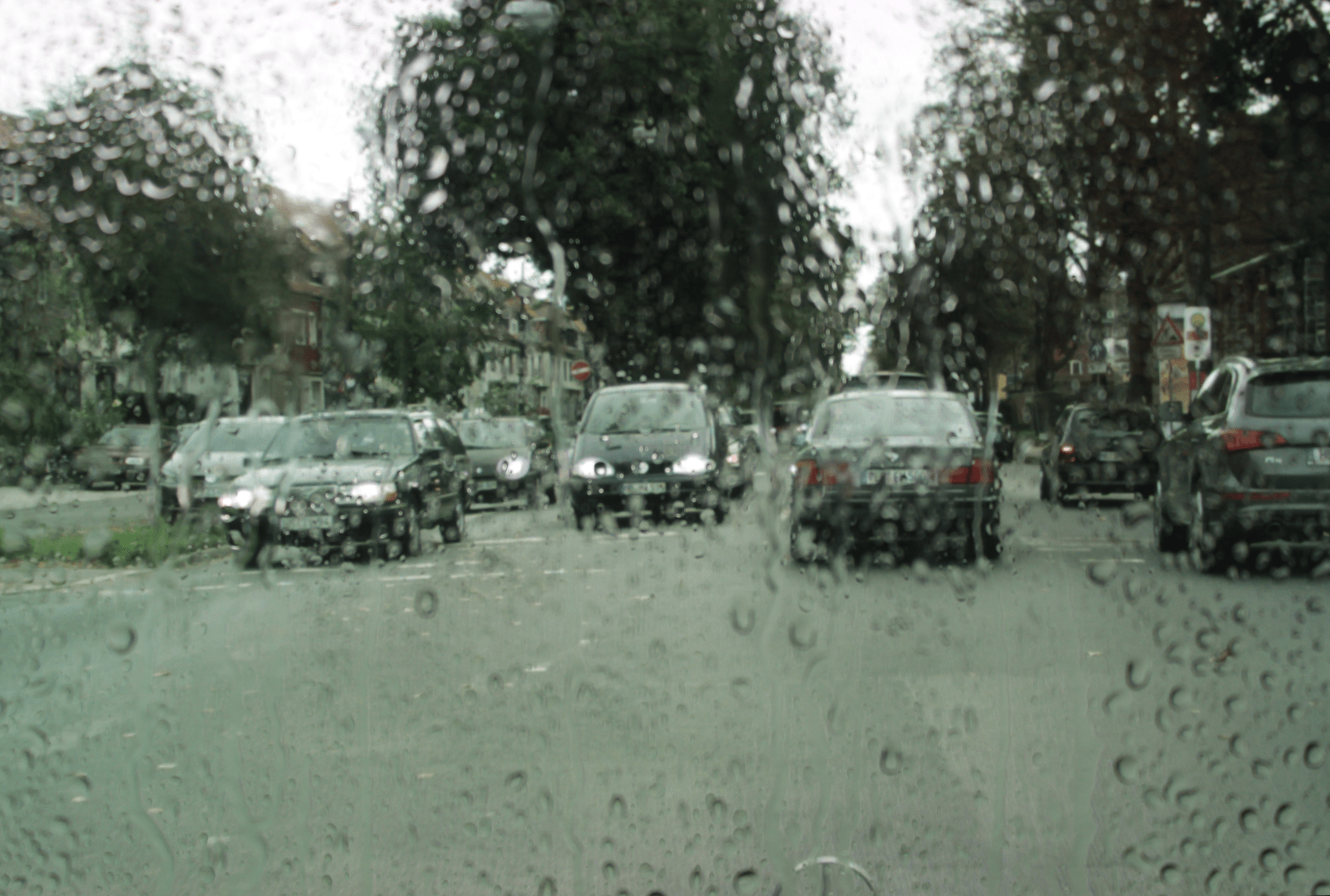}}%
\hfill 
{\includegraphics[width=0.33\textwidth]{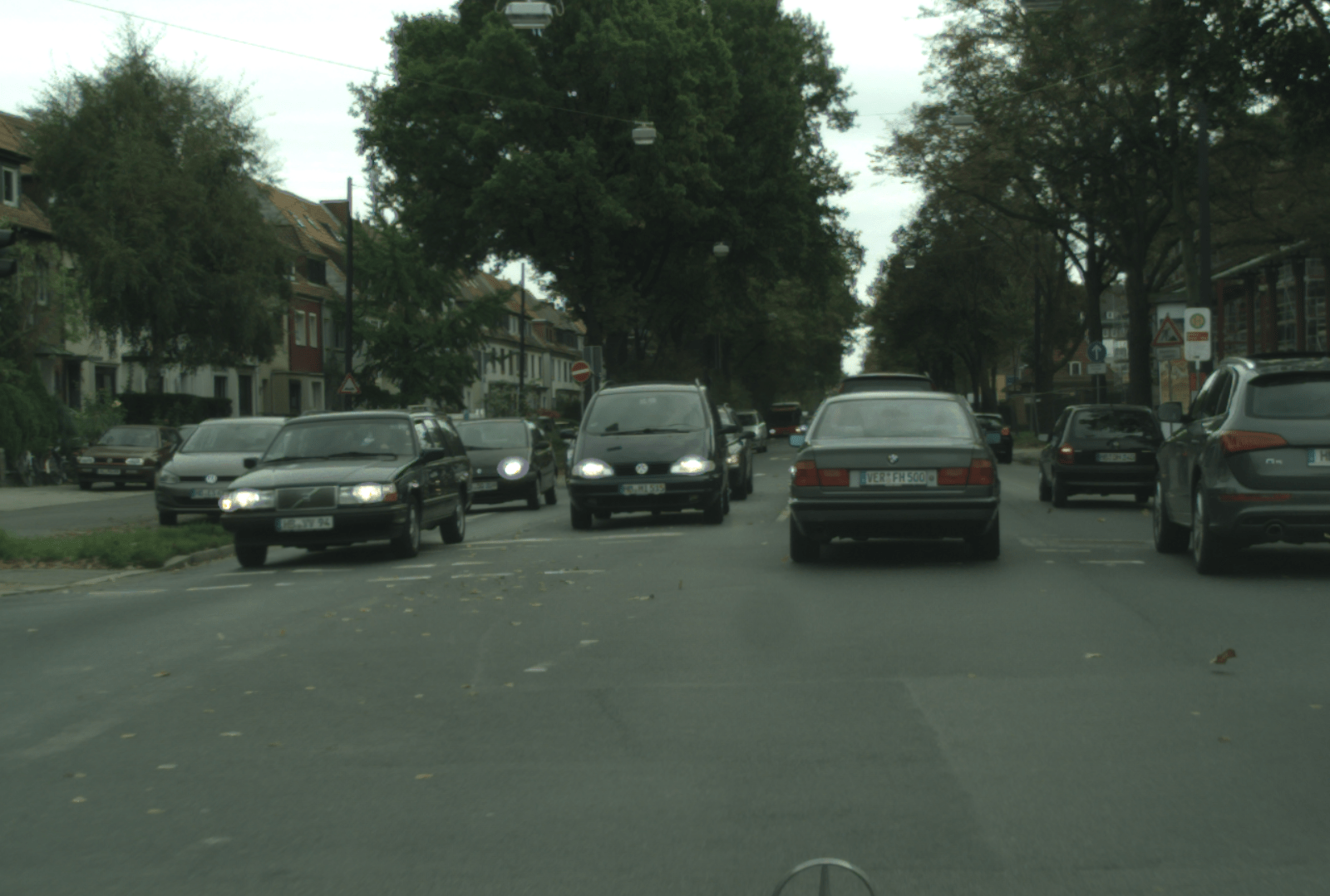}}%
\hfill 
{\includegraphics[width=0.33\textwidth]{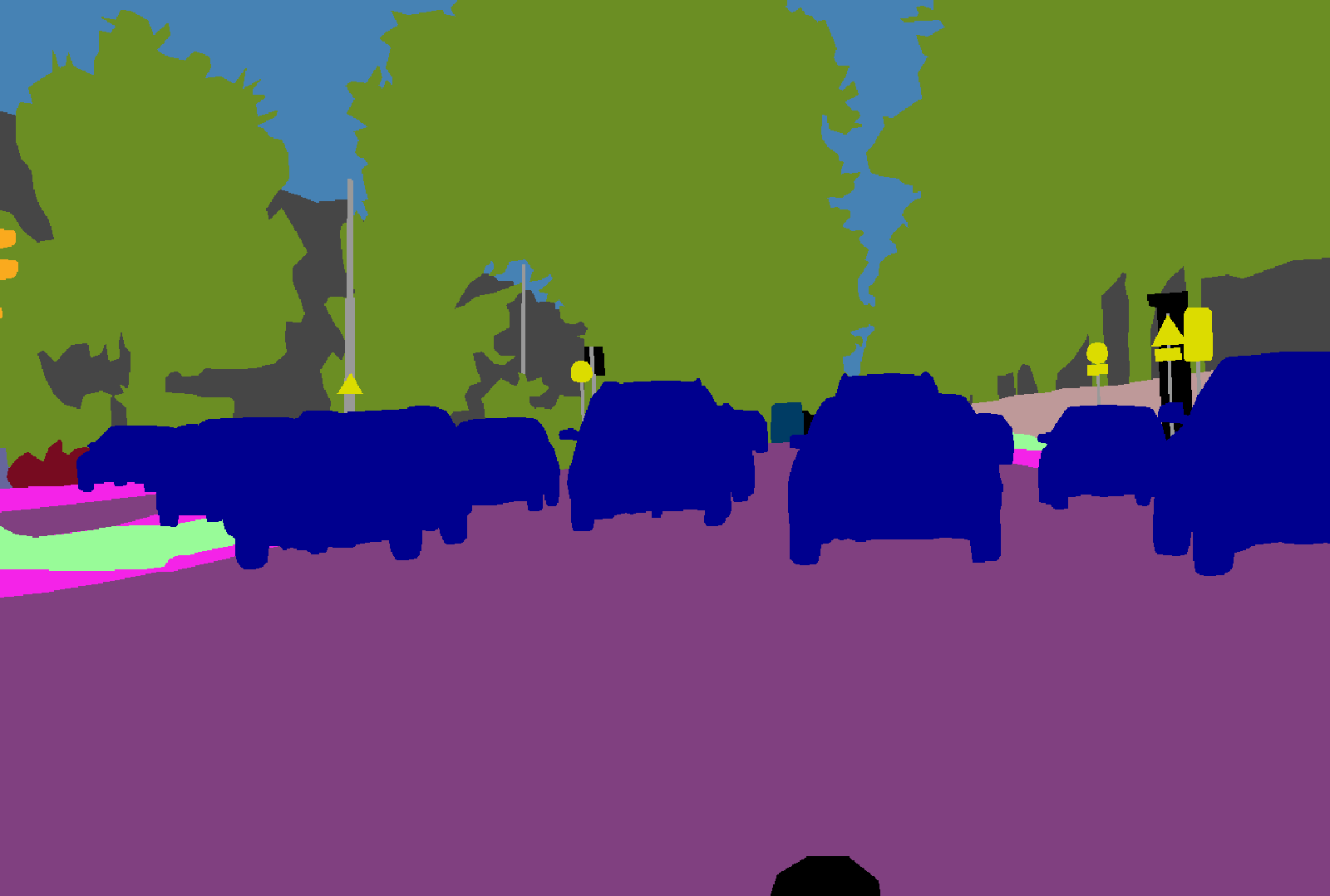}}%
\caption{Cityscapes-Rainy dataset. Left to right: Rainy, Clear ground-truth, Segmentation ground-truth.}
\label{out_of_sample}
\end{figure*}

We start by calibrating our camera for the desired zoom level, and roughly align it such that its optical axis is normal to the screen. For our particular experiment, the surface of the screen was 320mm from the outer lens, the glass pane was offset 170mm from the screen surface, and tilted to 20 degrees, simulating a windshield.
After setting the focus, we capture a test pattern that is used to estimate a homography between the monitor and the camera image. Since the image displayed on the screen already represents a 2D projection of the world, we have the advantage of being able to accurately align the camera image with the ground truth image using the estimated homography. 

The first pass of data acquisition involves capturing clear images through the glass pane, without any water, in order to account for the refraction of the glass. The second pass involves turning on the water nozzles (with a randomized pressure regimen) and capturing the same images again, resulting in corresponding rainy data. The data acquisition process must take place in a dark room to minimize stray reflections and changes in illumination. All data acquisition is done automatically and unattended, at approximately $1$ Hz, once the rig has been set up, and any misalignment is corrected using the computed homography. Note that the datasets are collected with the room lights OFF, without any natural or artificial lighting.

\subsection{The deraining model}

The principal purpose of our study is not to present a new architecture but to show that a high resolution monitor can be adequately used for data acquisition, hence we base our convolutional de-raining model on the widely-used architecture of Pix2PixHD \cite{wang2018highresolution}, with $4$ convolutional layers, 6 ResNet \cite{he2016deep} blocks and 4 up-convolutional layers. We add additive skip connections between each down- and up-convolutional layer, observing that a large part of the input structure, illumination and details can be kept and copied at the output. Similar to \cite{wang2018highresolution} and \cite{porav2019}, we do not make use of any unstructured pixel-wise losses, instead choosing a PatchGAN \cite{li2016precomputed} adversarial loss combined with discriminator feature losses and a VGG-based perceptual loss.

\subsection{Losses}

On the generator output, we apply an adversarial loss:
\begin{equation}
 \mathcal{L}_{adv}= (D(G(I_{\mathrm{rainy}}))-1)^2.
\end{equation}
With the discriminator being trained to minimize:
\begin{equation}
 \mathcal{L}_{disc}= (D(I_{\mathrm{clear}})-1)^2 + (D(I_{\mathrm{de-rained}}))^2,
\end{equation} 
A VGG-based perceptual loss \cite{Johnson2016Perceptual} is additionally used:
\begingroup
\setlength{\medmuskip}{0mu}
\setlength{\thinmuskip}{0mu}
\setlength{\thickmuskip}{0mu}
\setlength{\mathindent}{-0.1cm}
\begin{equation}
 \mathcal{L}_{\mathrm{perc}}\hspace{-0.2mm}=\hspace{-0.4mm}\sum_{i=1}^{n_{VGG}}\hspace{-0.5mm}\frac{1}{w_{i}^{perc}}{\lVert VGG(I_{\mathrm{clear}})_{i} - VGG(G(I_{\mathrm{rainy}}))_{i} \rVert}_{1}\hspace{-3mm}
\end{equation} with $n_{VGG}$ denoting the number of layers used in the loss and $w_{i}^{perc}=2^{(n_{VGG}-i)}$ weighing the contribution of each layer.
\endgroup
We also use a multi-scale discriminator feature loss  \cite{wang2018highresolution}:
\begin{equation}
 \mathcal{L}_{\mathrm{msadv}}=\sum_{i=1}^{n_{ADV}}\frac{1}{w_{i}^{adv}}{\lVert D(I_{\mathrm{clear}})_{i} - D(G(I_{\mathrm{rainy}}))_{i} \rVert}_{1},
\end{equation} with $n_{ADV}$ denoting the number of discriminator layers used in the loss and $w_{i}^{adv}=2^{(n_{ADV}-i)}$ weighing the contribution of each layer.

Finally, the full generator objective $\mathcal{L}_{\mathrm{gen}}$ is:
{\setlength{\mathindent}{-0.0cm}\begin{multline}
 \mathcal{L}_{\mathrm{gen}}\hspace{-0.5mm}=\hspace{-0.5mm}\lambda_{\mathrm{adv}}\hspace{-0.5mm}*\hspace{-0.5mm}\mathcal{L}_{\mathrm{adv}} + \lambda_{\mathrm{perc}}*\hspace{-0.5mm}\mathcal{L}_{\mathrm{perc}} + \lambda_{\mathrm{msadv}} * \hspace{-0.5mm}\mathcal{L}_{\mathrm{msadv}}\hspace{-1mm}
\end{multline}}
The hyperparameters $\lambda$ modulate the importance of the individual terms of the loss. We estimate a discriminator $D$ and generator $G$ such that:
\begin{equation}
 G, D = \underset{G,D}{\arg\min} \mathcal{L}_{\mathrm{gen}} + \mathcal{L}_{\mathrm{disc}}.
\end{equation}
\begin{table*}[!htbp]
	\begin{center}
		\begin{tabular}{@{} |c|c| C{0.6cm}*{1}{C{0.6cm}} C{0.6cm} *{1}{C{0.6cm}} @{}}
			\hline
			\multirow{3}*{\bf Dataset} &
			\multirow{3}*{\bf Model (ours) trained on} & \multicolumn{2}{c|}{\bf RAINY} & \multicolumn{2}{c|}{\bf DERAINED} \\
			& & PSNR&\multicolumn{1}{C{0.7cm}|}{SSIM}& PSNR &\multicolumn{1}{C{0.7cm}|}{SSIM} \\ \hline
			Cityscapes-Rainy & Cityscapes-Original labels & 18.20&\multicolumn{1}{c|}{0.6865}& \textbf{25.76}&\multicolumn{1}{c|}{\textbf{0.8817}} \\ \hline
			Cityscapes-Rainy & Cityscapes-Photographed labels & 18.20&\multicolumn{1}{c|}{0.6865}& \textbf{21.20}&\multicolumn{1}{c|}{\textbf{0.8294}} \\ \hline
			BDD-Rainy & Cityscapes-Original labels + BDD-Original labels & 17.84&\multicolumn{1}{c|}{0.6500}& \textbf{24.02}&\multicolumn{1}{c|}{\textbf{0.8587}} \\ \hline
		\end{tabular}
		\caption{Reconstruction results}
	\end{center}
	\label{Tab:recresults}
\end{table*}

\subsection{Training}
We employ a similar regimen to \cite{porav2019}, training all models for 100 epochs, using the Adam optimizer, with a base learning rate of $0.0002$. 
\section{Results}
We collect rainy versions of Cityscapes\cite{Cordts2016Cityscapes} and BDD\cite{bdd} datasets, and study reconstruction and segmentation quality. 

\subsection{Reconstruction results}
Image reconstruction results are presented in Table \textcolor{red}{1}. We train three different models, one using the original Cityscapes clear images as ground-truth (\textbf{Cityscapes-Original}), one using the photographed Cityscapes clear images as ground-truth (\textbf{Cityscapes-Photograph}), and one trained on both the the original Cityscapes and original BDD clear images. We present the deraining performance of these three models on both Cityscapes-Rainy and BDD-Rainy. All models significantly improve the quality of the rain-affected images. Secondly, Table \textcolor{red}{2} shows the performance of our model on a third-party dataset by Qian et al.\cite{qian2017}, containing 861 training images and 59 testing images, acquired using a glass-pane manually sprayed with water and a DSLR camera in an outdoor environment. On one hand, we show that our model architecture achieves state-of-the-art results when trained on the target dataset. On the other hand, training solely on Cityscapes-Rainy does not produce top results. We observe that the model performs adequately on droplets that are in-focus or nearly in-focus, but inadequately on diffuse droplets, since these were not present in our training data. However, we show that finetuning (\textbf{Ours - trained on Cityscapes-Rainy, finetuned on Qian-sample(112 img.)}) the model with a very small random sample of the target training set (approximately 10\%) yields excellent performance, showing that pre-training a model on our automatically-collected data significantly reduces the need for labelled target domain data.
\begin{table}[!htbp]
	\begin{center}
		\begin{tabular}{@{}|C{5.3cm}|C{0.3cm} C{0.3cm}@{}}
			\hline
			\multirow{2}*{\bf Model vs. Dataset} & \multicolumn{2}{c|}{\bf Dataset of \cite{qian2017}} \\
			& \multicolumn{1}{c}{PSNR}&\multicolumn{1}{c|}{SSIM} \\ \hline
			RAW & \multicolumn{1}{c|}{24.09}&\multicolumn{1}{c|}{0.8518} \\ \hline
			Eigen13\cite{eigen2013} & \multicolumn{1}{c|}{28.59}&\multicolumn{1}{c|}{0.6726} \\ \hline
			Pix2Pix\cite{pix2pix} & \multicolumn{1}{c|}{30.14}&\multicolumn{1}{c|}{0.8299} \\ \hline
			Qian et al.(no att.)\cite{qian2017} & \multicolumn{1}{c|}{30.88}&\multicolumn{1}{c|}{0.8670} \\ \hline
			Qian et al.(full att.)\cite{qian2017} & \multicolumn{1}{c|}{31.51}&\multicolumn{1}{c|}{\textbf{0.9213}} \\ \hline
			Ours - trained on Qian-Full \\(862 img.) & \multicolumn{1}{c|}{\textbf{31.55}}&\multicolumn{1}{c|}{0.9020} \\ \hline
			Ours - trained on Cityscapes-Rainy & \multicolumn{1}{c|}{27.52}&\multicolumn{1}{c|}{0.8716} \\ \hline
			Ours - trained on Cityscapes-Rainy, finetuned on Qian-sample(112 img.) & \multicolumn{1}{c|}{30.21}&\multicolumn{1}{c|}{0.8953} \\ \hline
		\end{tabular}
		\caption{Reconstruction quality comparison to state of the art}
	\end{center}
	\label{tab:qian}
\end{table}

\subsection{Segmentation results}
\begin{figure}[!htbp]
\centering
{\includegraphics[width=0.495\columnwidth]{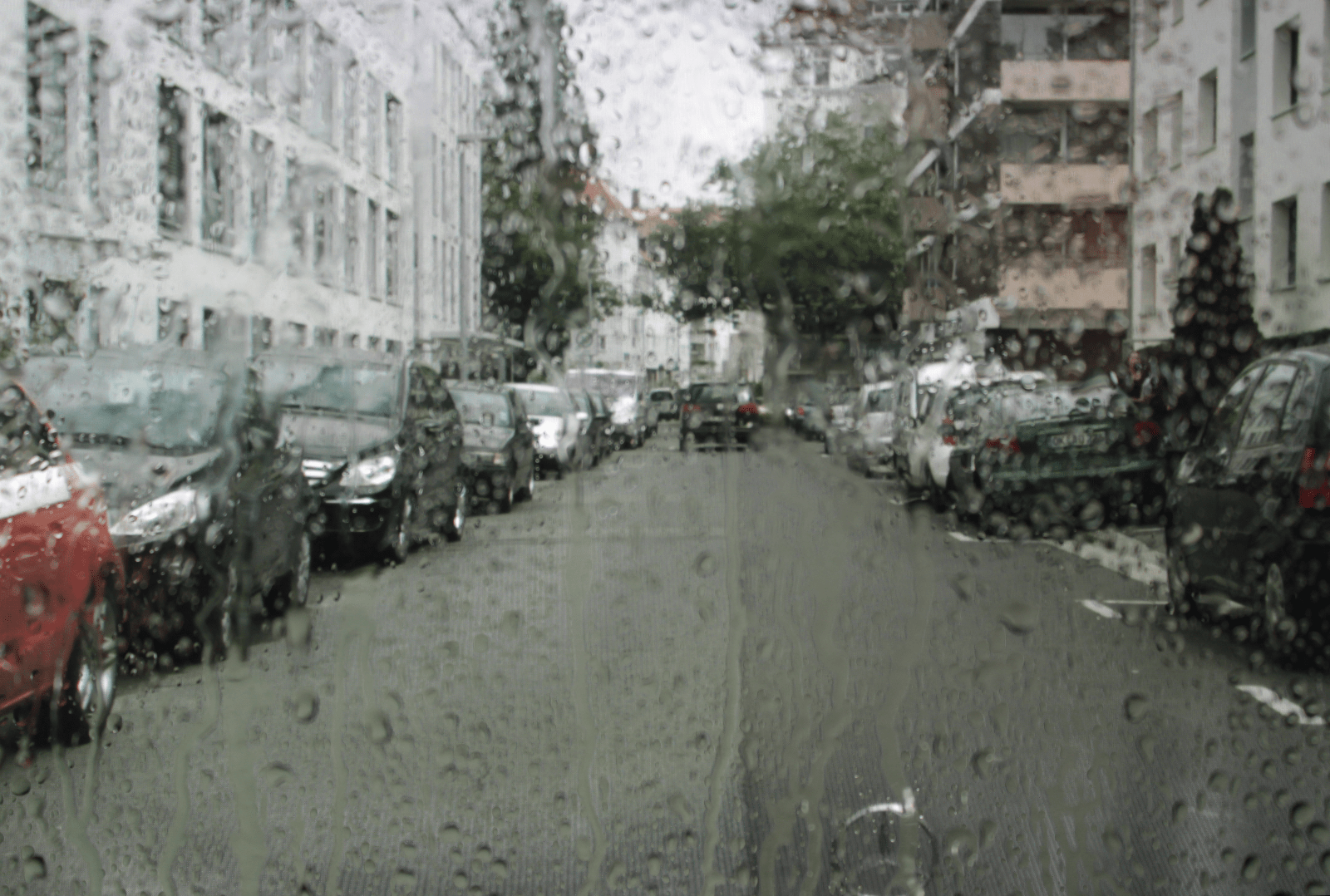}}%
\hfill 
{\includegraphics[width=0.495\columnwidth]{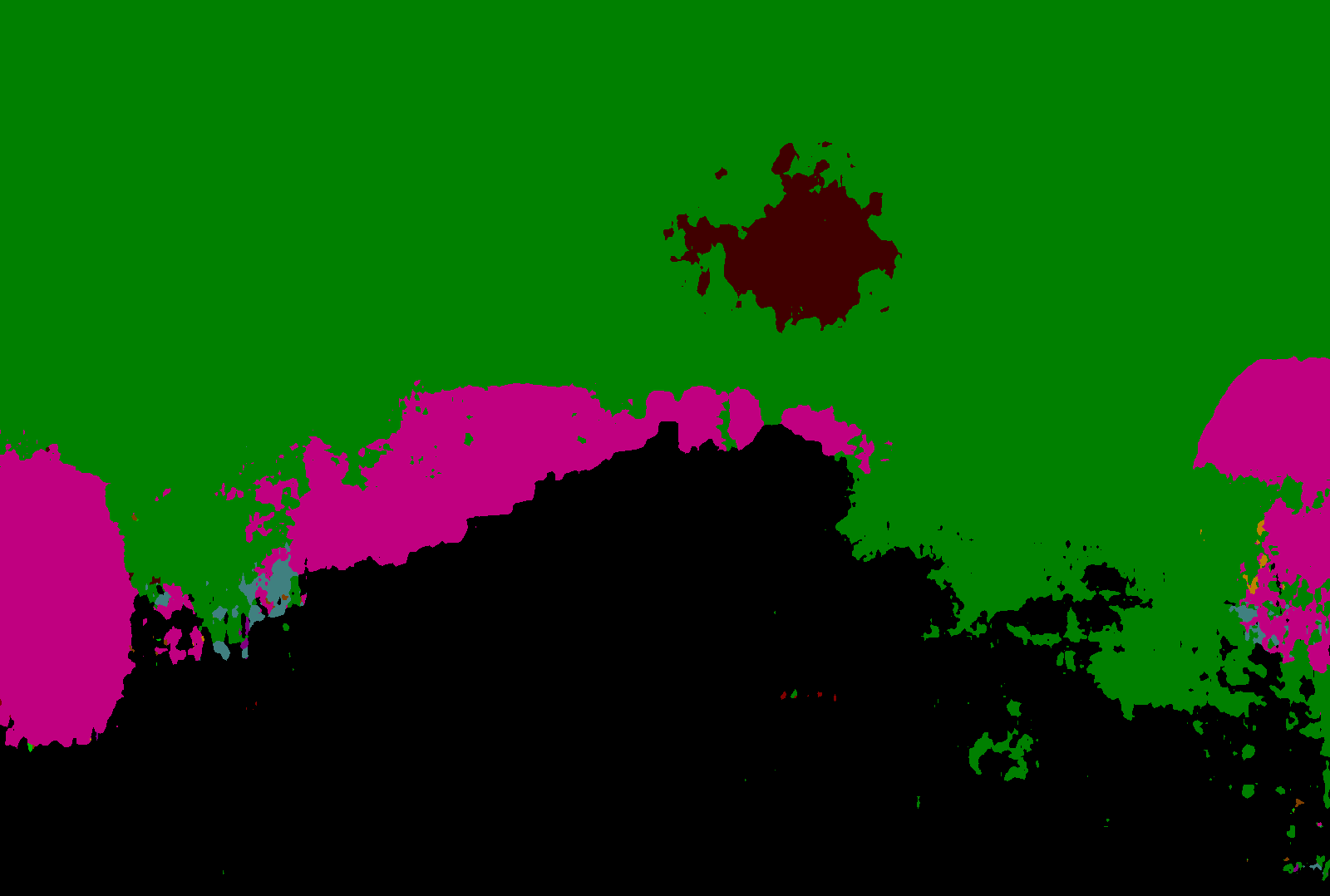}}%
\\ 
{\includegraphics[width=0.495\columnwidth]{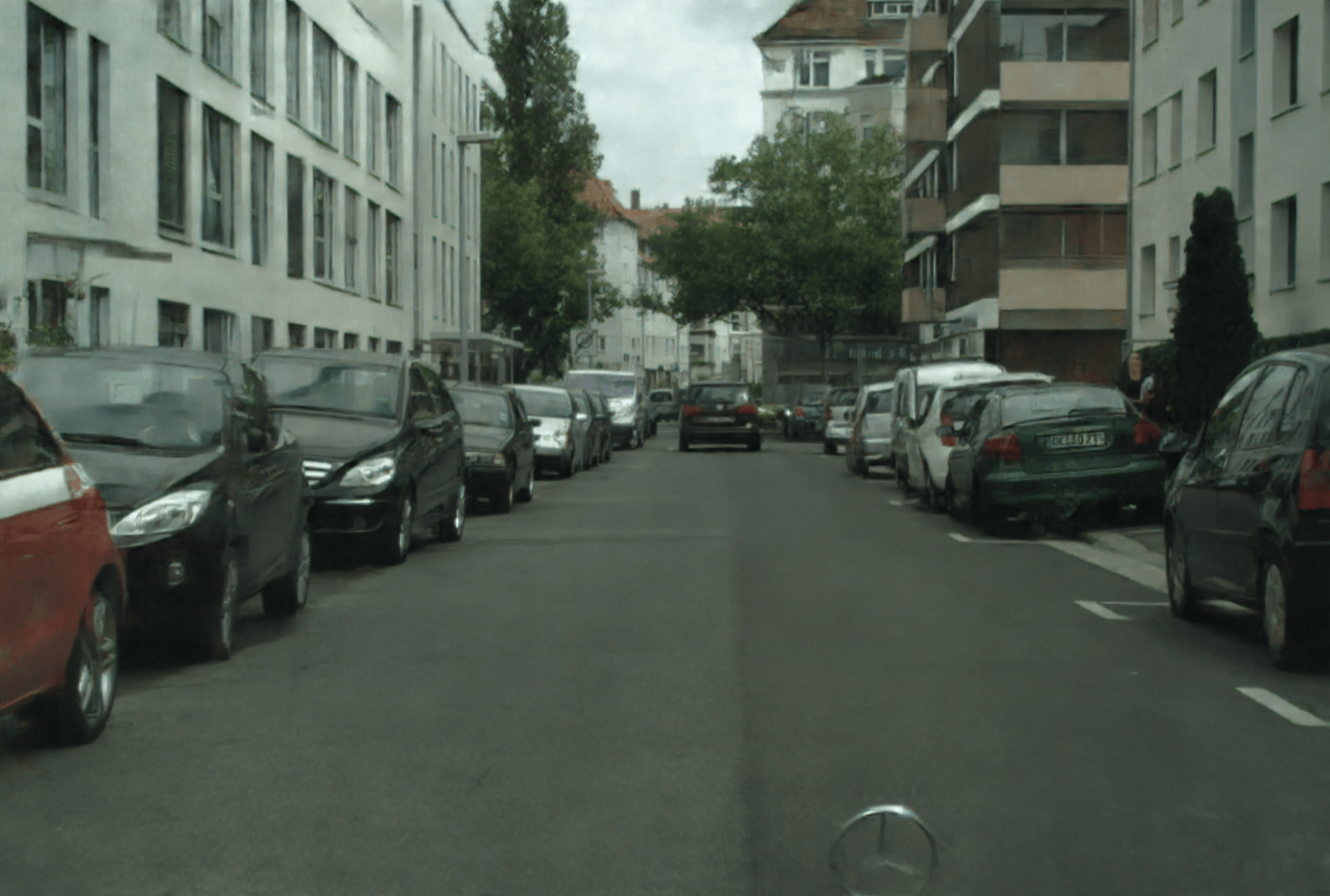}}%
\hfill 
{\includegraphics[width=0.495\columnwidth]{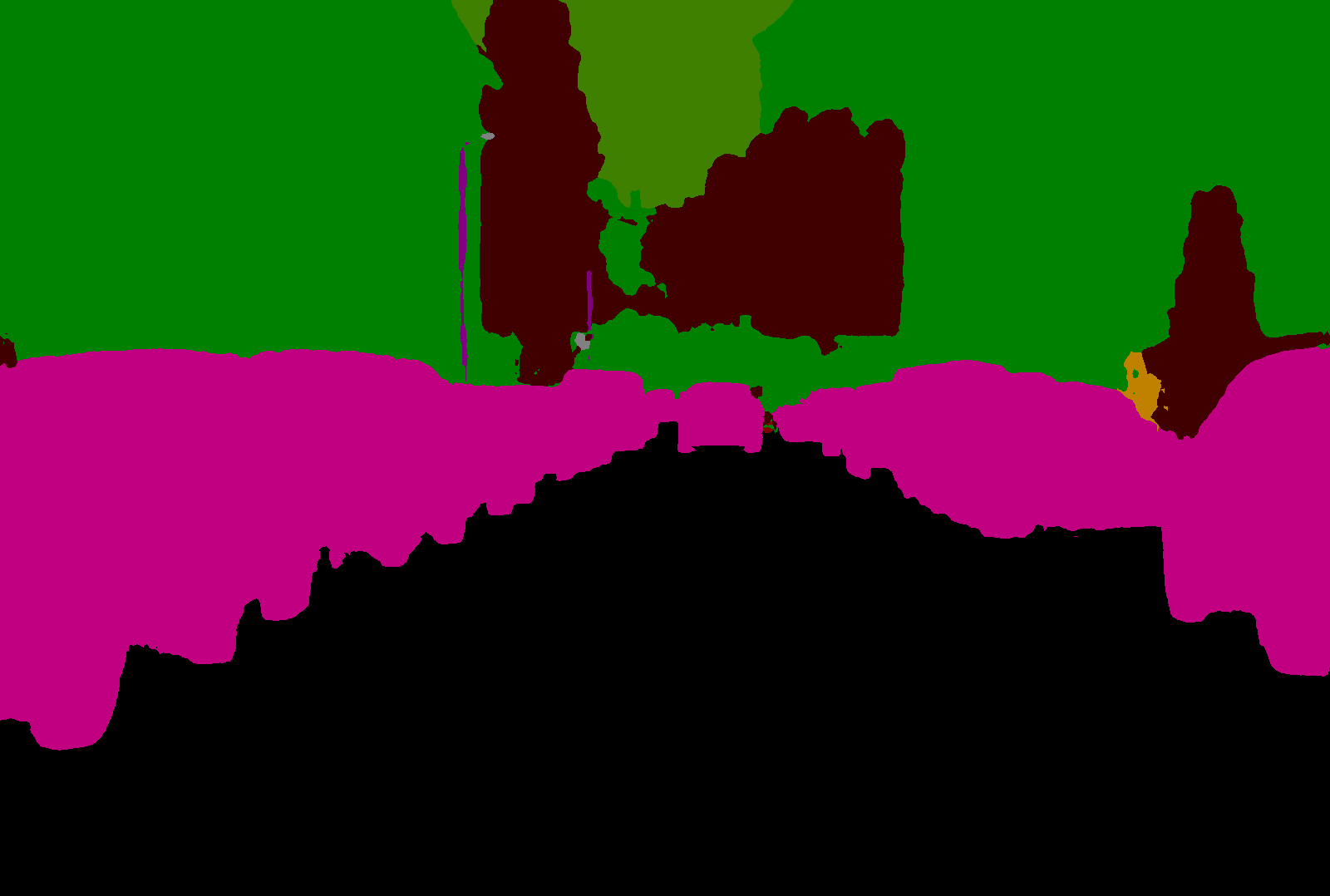}}%
\caption{Segmentation results. Top: rainy image. Bottom: derained image.}
\label{segmentation_results}
\end{figure}

We show that rainy images severely degrade semantic segmentation performance, and that deraining the images \textbf{restores} performance. We use an off-the shelf segmentation model (DeepLab v3+\cite{deeplabv3plus}, trained on the original, vanilla Cityscapes dataset), and derain Cityscapes-Rainy images using the model with top-performing reconstruction results (results in Table \textcolor{red}{3} and Fig. \ref{segmentation_results}). The segmentation model achieves an mIOU of 0.79 on the original Cityscapes dataset, 0.71 on photographed (dry) Cityscapes, 0.17 on rainy Cityscapes, and a significantly improved value of 0.63 on rainy images that were derained using our model.  Additionally, we show in the first two rows of Table \textcolor{red}{3} that taking photographs of a high-resolution monitor does not lead to a large loss in segmentation quality, as compared to the original images. We believe that the difference in performance (mIOU of 0.71 vs. 0.79 for the original images) can be reduced further by using a higher-resolution setup, better lens and a better alignment methodology.
\begin{table}[!htbp]
	\begin{center}
		\begin{tabular}{@{} |c| C{0.8cm} @{}}
			\hline
			\multirow{1}*{\bf Cityscapes Img. vs. Segm. Model}
			& \multicolumn{1}{C{0.8cm}|}{mIoU} \\ \hline
			CLEAR-Original & \multicolumn{1}{c|}{0.7901}\\ \hline
			CLEAR-Photographed & \multicolumn{1}{c|}{0.7137}\\ \hline
			RAINY & \multicolumn{1}{c|}{0.1776} \\ \hline
			DERAINED & \multicolumn{1}{c|}{\textbf{0.6327}} \\ \hline
		\end{tabular}
		\caption{Cityscapes Semantic segmentation results}
	\end{center}
	\label{tab:semseg}
\end{table}


\subsection{Qualitative results}
\begin{figure}[!htbp]
\centering
{\includegraphics[width=0.495\columnwidth]{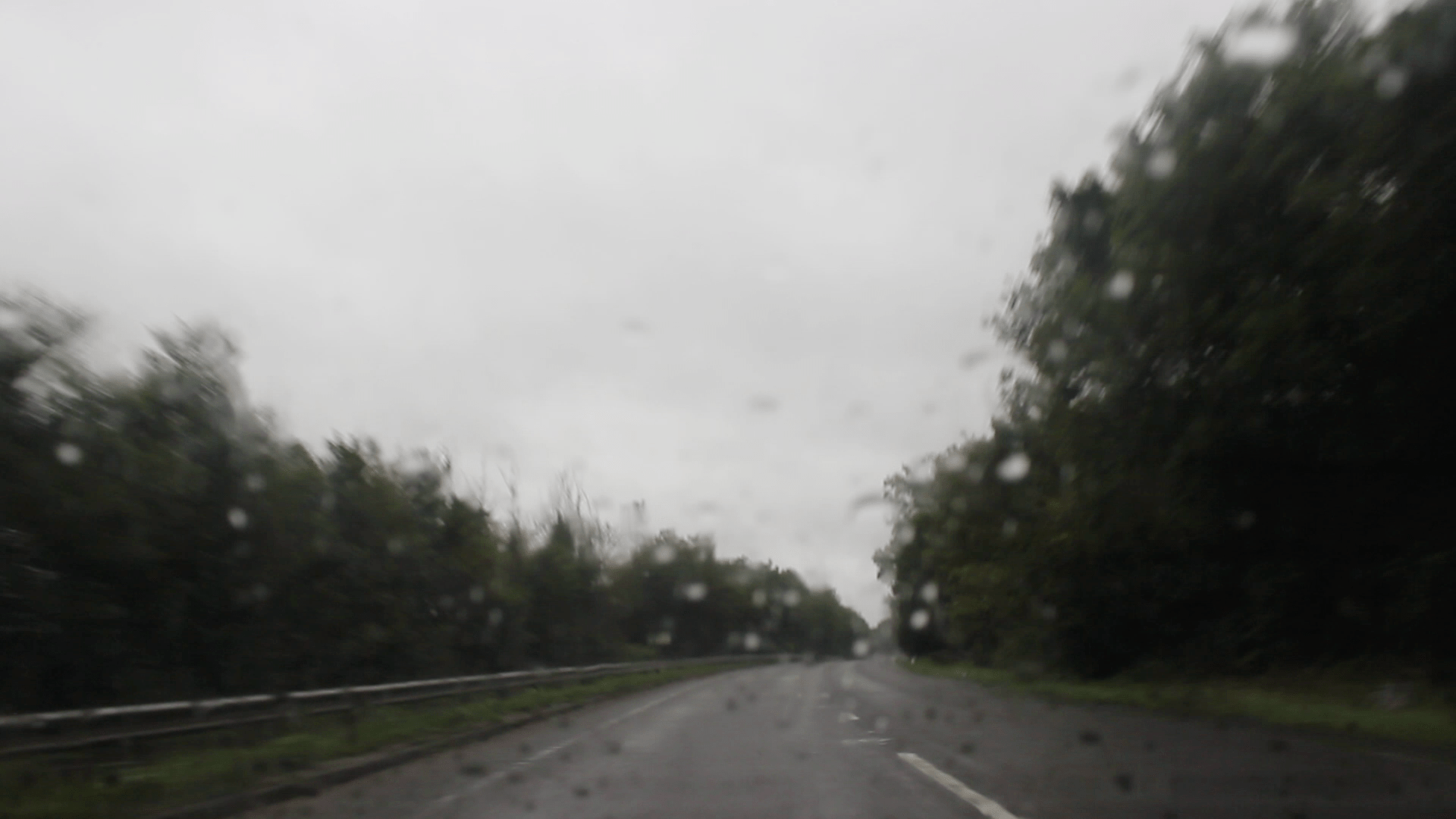}}%
\hfill 
{\includegraphics[width=0.495\columnwidth]{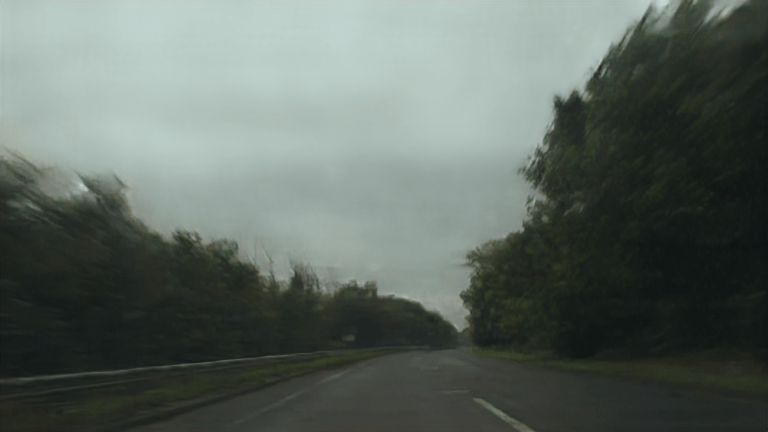}}%
\caption{Qualitative results on \textbf{unseen} real-rain domain captured using a smartphone camera, from inside a moving vehicle. }
\label{qualitative_results}
\end{figure}

We show that our model generalises to other camera setups by recording a real-rainy dataset from inside a moving vehicle, using a camera from a mid-level smartphone. We observe satisfactory reconstruction performance for varying droplet and splatter shapes. We present an example of de-raining of images from such an \textbf{unseen} real domain in Figure \ref{qualitative_results}. Outside of quantitative results on third-party rain datasets presented in Table \textcolor{red}{2} and on our own data presented in Table \textcolor{red}{1}, we are severely limited in offering further numerical analyses by the lack of third-party rainy datasets with either a clear groundtruth or semantic segmentation labels.
\section{Conclusion}
We have shown that a very simple automated setup - a high resolution computer screen, a glass pane sprayed with water nozzles, and a camera - is suitable for generating rainy data from arbitrary datasets, and more importantly from those with ground truth for auxiliary tasks such as semantic segmentation. Both our quantitative and qualitative results show that data collected in this fashion can be effectively used to train image de-raining and de-noising models that generalise well to other environments, and that this data is valuable for closing domain gaps via fine-tuning, especially when the target data is hard to acquire.

Currently, the system is able to tackle adherent rain, while other effects such as atmospheric misting or fog are out of the scope of this work. Attempts at reducing the impact of these other effects can be found throughout the literature, with a good example being \cite{vangoolfoggy}. Additionally, while the current implementation involves a windscreen-like setup where windscreen wipers could be envisioned to work and alleviate some of the issues stemming from adherent rain, there are many other camera/sensor locations where wipers would be impractical and which would benefit from our approach. Finally, even if effects such as taillight reflections on the wet ground cannot be captured or introduced, this does not reduce the usefulness nor the validity of datasets obtained using this method when used for image reconstruction and denoising purposes. While such a dataset does not capture the full effects of rain, it can be used, at a minimum, to massively and effectively reduce the amount (and cost) of labelled/ground-truth real rainy data that needs to be collected.
\begin{figure}[!tbph]
\centering
{\includegraphics[width=0.496\columnwidth,height=2.32cm]{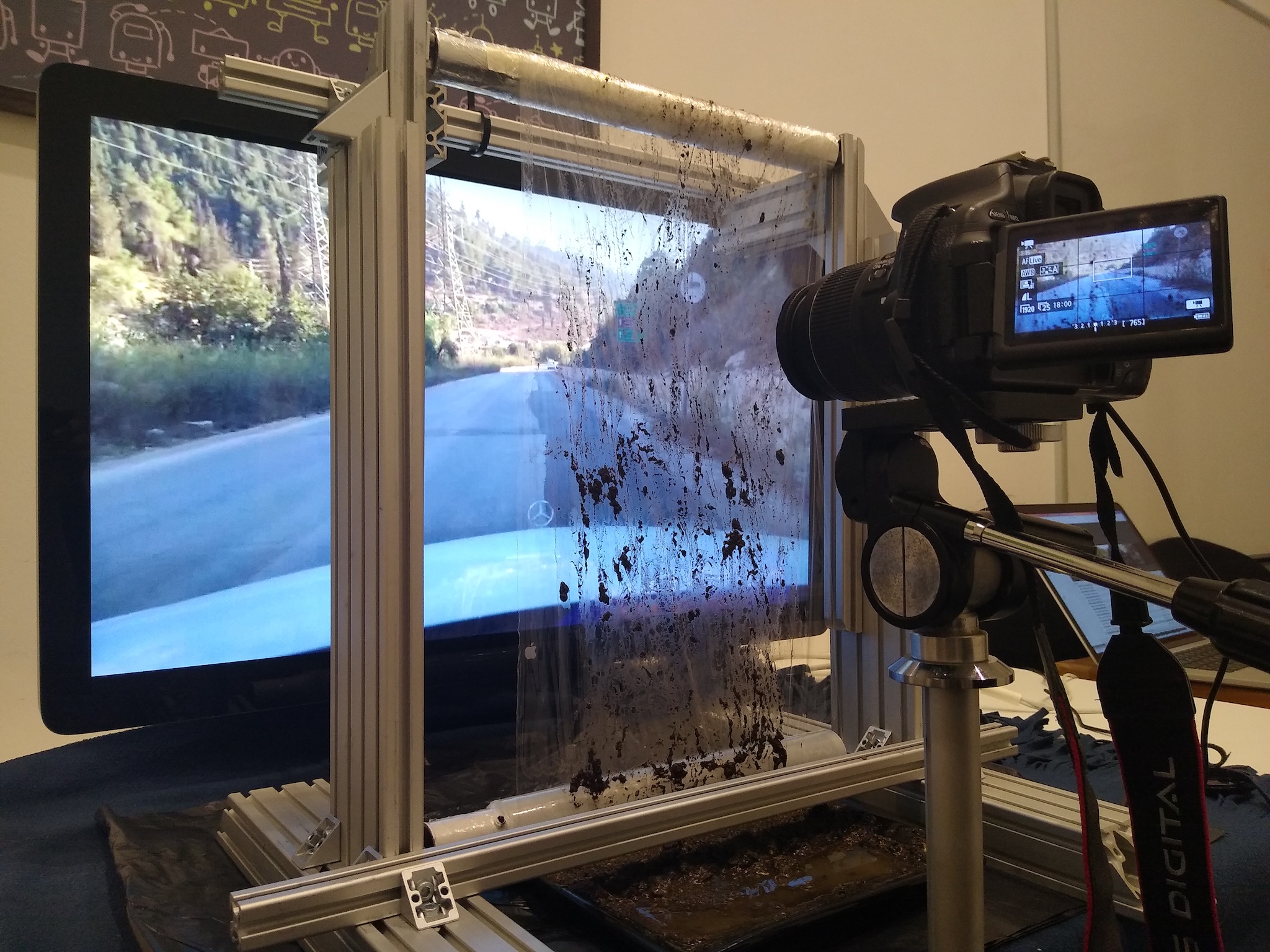}}%
\hfill 
{\includegraphics[width=0.496\columnwidth]{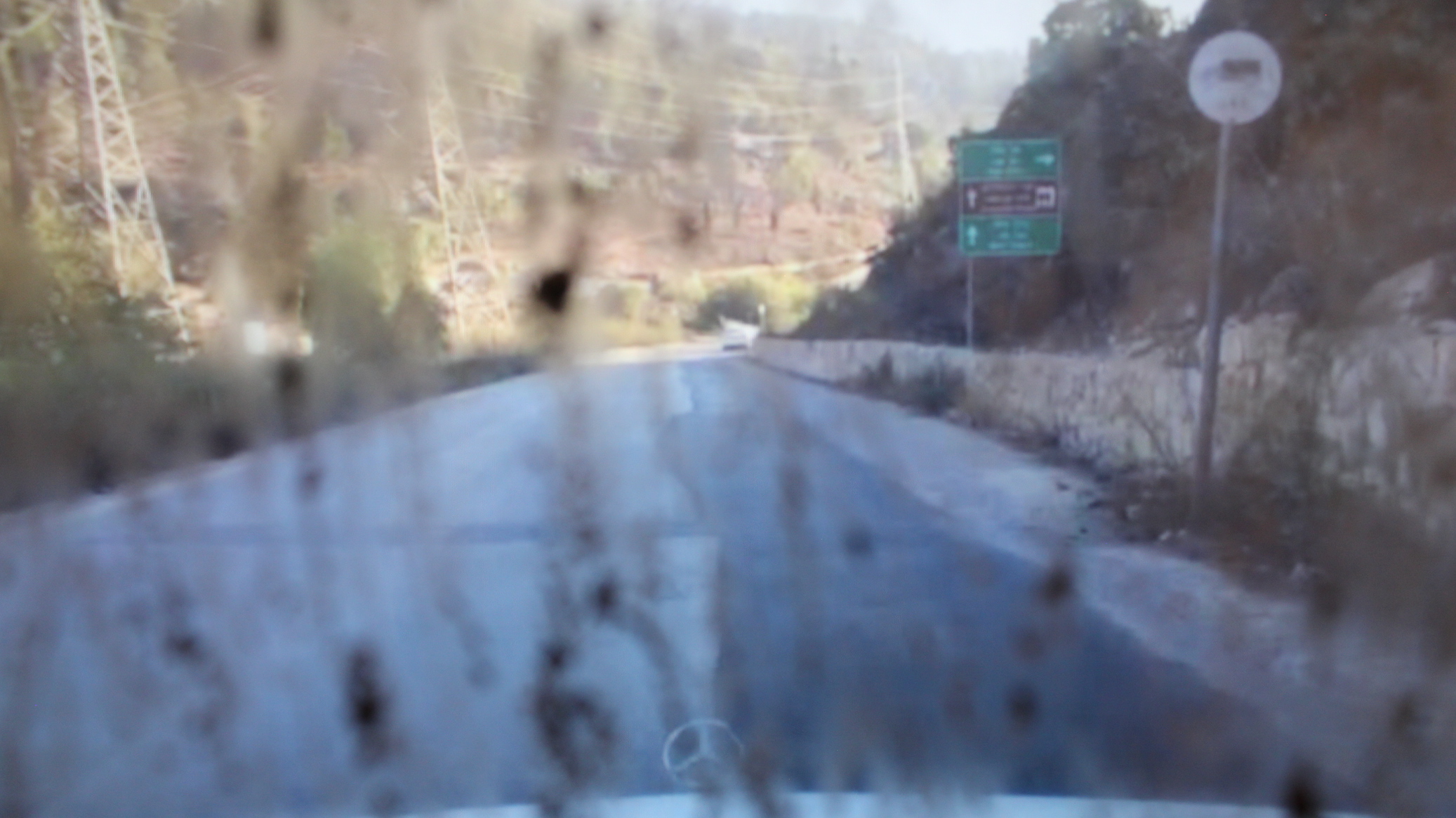}}
\caption{Left: proof of concept. Right: muddy image.}
\label{Fig:dirt}
\end{figure}

\section{Limitations and future work}
The proposed setup is only able to simulate adherent contaminants. For example, wind, heavy snow, hailstorm and other atmospheric contaminants cannot be simulated if they are supposed to change the structure and appearance of objects already captured in the image.
Additionally, the ease of generating data with this device comes at the cost of losing some image quality (see Table \textcolor{red}{3}, second row).
Another adherent contaminant that creates an obscuring effect is dirt (dust, mud etc). Soil is synthetically generated in \cite{uricar2019}, whereas we propose as future work to use a rolling clear film to gather real mud (see Fig. \ref{Fig:dirt}).


{\small
\bibliographystyle{ieee}
\bibliography{references}
}

\end{document}